\documentclass{article}
\usepackage{cite}
\usepackage{amsmath,amssymb,amsfonts}
\usepackage{algorithmic}
\usepackage{graphicx}
\usepackage{textcomp}
\usepackage{xcolor}
\usepackage{hyperref}
\usepackage{soul} 
\usepackage{color}  
\usepackage{indentfirst}
\usepackage{multirow}
\usepackage{mathtools}
\usepackage{hhline}

\usepackage{color, colortbl}

\def\robotname{SKOOTR}
\def\deepblue{\href{https://deepblue.lib.umich.edu/data/concern/data_sets/gq67js03v/anonymous_link/3d2b251d202af03ccda4e0dcfa778e89a2b111bc898eb8880305376ac456a60c
}{Deep Blue Data Repository}}

\def\BibTeX{{\rm B\kern-.05em{\sc i\kern-.025em b}\kern-.08em
    T\kern-.1667em\lower.7ex\hbox{E}\kern-.125emX}}
\begin{document}

\title{\robotname: A SKating, Omni-Oriented, Tripedal Robot}

\author{Adam Joshua Hung$^{1}$, 
Challen Enninful Adu$^{1}$, and 
Talia Y. Moore$^{1,2}$
\thanks{*This work was not supported by any organization.}
\thanks{$^{1}$Dept. of Robotics, University of Michigan, Ann Arbor, MI, USA
      }%
\thanks{$^{2}$Dept. of Mechanical Engineering, Ecology and Evolutionary Biology, Museum of Zoology, University of Michigan, Ann Arbor, MI, USA
      {\tt\small taliaym@umich.edu}}%
}

\maketitle

\begin{abstract}
In both animals and robots, locomotion capabilities are determined by the physical structure of the system.
The majority of legged animals and robots are bilaterally symmetric, which facilitates locomotion with consistent headings and obstacle traversal, but leads to constraints in their turning ability.
On the other hand, radially symmetric animals have demonstrated rapid turning abilities enabled by their omni-directional body plans.
Radially symmetric tripedal robots are able to turn instantaneously, but are commonly constrained by needing to change direction with every step, resulting in inefficient and less stable locomotion.
We address these challenges by introducing a novel design for a tripedal robot that has both frictional and rolling contacts.
Additionally, a freely rotating central sphere provides an added contact point so the robot can retain a stable tripod base of support while lifting and pushing with any one of its legs.
The SKating, Omni-Oriented, Tripedal Robot (SKOOTR) is more versatile and stable than other existing tripedal robots. 
It is capable of multiple forward gaits, multiple turning maneuvers, obstacle traversal, and stair climbing.
\robotname\ has been designed to facilitate customization for diverse applications: it is fully open-source, is constructed with 3D printed or off-the-shelf parts, and costs approximately \$500 USD to build.
\end{abstract}


\section{Introduction}\label{sec:intro}

The configuration of a robot's actuators and linkages determine its set of possible gaits or locomotion capabilities \cite{yosinski2011evolving,zykov2004evolving}. 
The legged robot form factors and gaits often draw inspiration from biology \cite{iida2016biologically}. 
As vertebrates ourselves, roboticists tend to draw inspiration from other vertebrates (e.g., dogs, cheetahs, turtles, and goats), resulting in a bilaterally symmetric body plan for most legged robots.
One of the main challenges of bilaterally symmetric body forms is rapid turning.
Even in animals, morphological and neurological specializations are required to overcome the constraints of a bilaterally symmetric body form to rapidly reorient and strike prey \cite{zeng_biomechanics_2018,haagensen2022exploring}.

\begin{figure}[h]
    \includegraphics[width=\columnwidth]{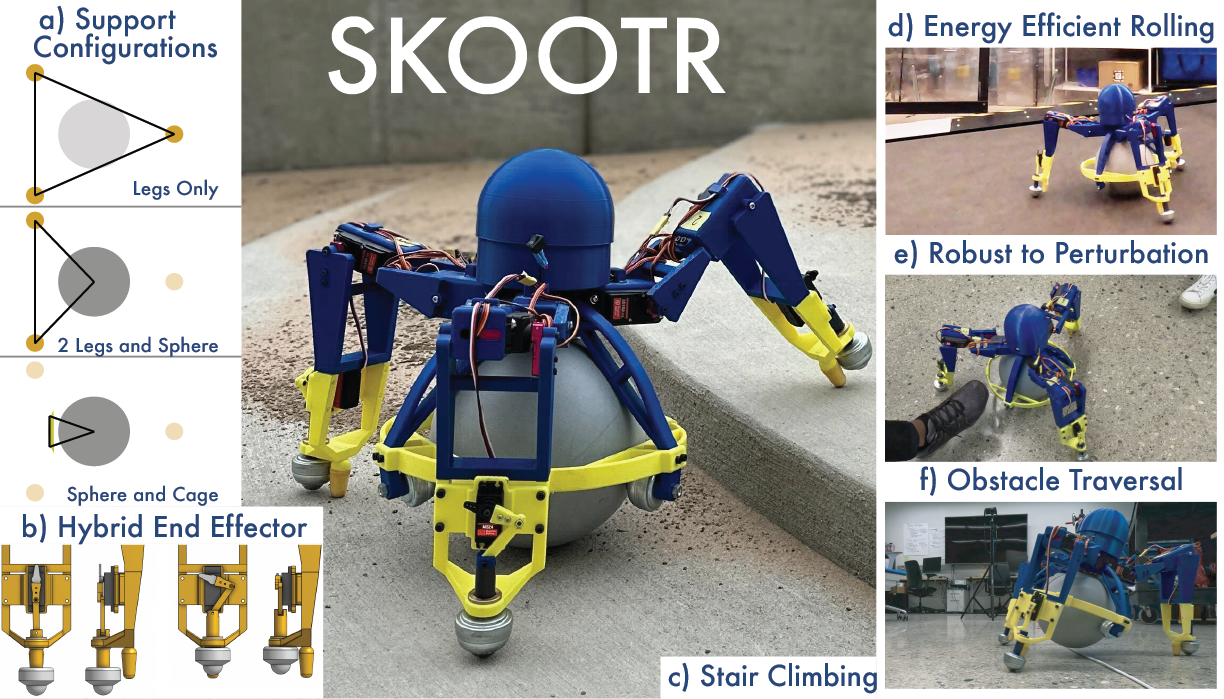}
    \caption{Overview of the specifications and capabilities of \robotname.
    \robotname\ has three different support modes:
    all feet, two feet and the center sphere, or the center sphere and the edge of the surrounding cage.
    Each foot has two contact modes: rolling or frictional.
    \robotname\ can successfully perform several tasks often associated with either rolling or legged robots.
    Like rolling robots, it is capable of energy efficient rolling down inclines or on level surfaces with external perturbations.
    Like legged robots, it is capable of obstacle traversal and stair climbing.}
    \label{fig:overview}
    \centering
    \vspace{-.2cm}
\end{figure}

In comparison, the methods of locomotion of radially symmetric animals are less well understood, but turning appears to be a common advantage of this body form. 
Radially symmetric brittle stars \cite{astley_getting_2012} and various cephalopods \cite{huffard_locomotion_2006, jastrebsky_kinematics_2015} frequently choose a different portion of their bodies as the ``front'' and adjust the motions of their limbs to move in that direction without rotating the body.
Thus, they can ''turn'' by instantaneously altering their heading (direction of velocity vector) without adjusting their body orientation.

Radially symmetric robots have previously been designed, but few leverage the omni-directional maneuverability of their biological counterparts \cite{ishikawa2012tripedal,kim2019simple}.
Tripedal configurations implement the minimum number of legs required for locomotion with radial symmetry. 
For example, the tripedal robot STriDER \cite{heaston_strider_2007} lifts one leg and swings it under the center of mass and across the line defined by the two other legs.
Another tripedal robot, Rotopod \cite{lyons_rotopod_2007}, uses a rotating reaction mass to pivot two legs about a third leg to take each step.
In these robots, step direction is constrained by the tripedal form factor, resulting in each step conferring a 120\textdegree\ change in heading.
While this facilitates rapid turning, these tripedal robots require inefficient lateral motion of the center of mass to travel in a consistent direction.
Furthermore, the gaits of these tripedal robots are less stable than their quadrupedal, hexapedal, or wheeled counterparts due to only having two points of contact with the ground during motion.

\begin{figure*}[t]
    \centering
    \includegraphics[width=\textwidth]{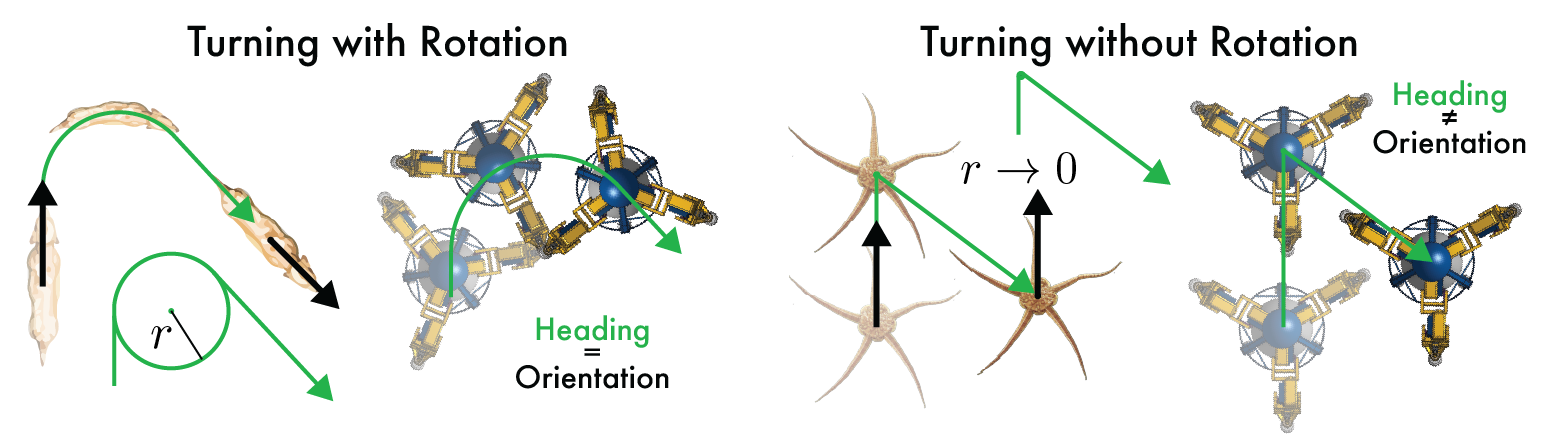}
    \caption{Two different turning maneuvers.
    Turning with rotation keeps the heading (in green) and orientation (in black) aligned throughout the maneuver, as demonstrated by the top-down view of a dog turning.
    Turning without rotation maintains the same orientation of the body throughout the maneuver, but changes heading.
    This is demonstrated by a brittle star that simply selects a different leading leg to initiate a change in heading.
    The turning radius, $r$, is defined by the radius of the largest circle that can be inscribed in the turn\cite{howland1974optimal}.
    Turning with rotation has a much larger $r$ than turning without rotation.}
    \label{fig:turn}
    \vspace{-.2cm}
\end{figure*}

Alternatively, an aquatic, sea star-inspired soft robot demonstrates how radial symmetry can enhance turning ability in water \cite{huang2021numerical}.
Because this soft robot is buoyant, it is not susceptible to toppling over as it moves along the planar surface of the water, and can therefore maintain a continuous heading by performing consecutive accelerations in the same direction.
Like the brittle star, it can also instantaneously change heading to perform a rapid turning maneuver by selecting a different side of the body to lead the motion.

To enhance stability, many robots either roll on wheels that never leave contact with the substrate, or adopt a multi-legged configuration that confers greater stability as leg number increases.
Some robots combine rolling and legged configurations, with the goal of enhancing energy efficiency while maintaining  environmental traversability \cite{saranli_design_2000, allen_abstracted_2003, bjelonic_keep_2019}.
However, the vast majority of these robots use single-axis wheels for contact, thereby constraining the minimum turning radius.
Omni-wheels have been invented to address this constraint, but typical robots outfitted with omni-wheels are unable to surmount even small obstacles or stairs \cite{pin1994new,wu2008novel}.


A spherical rolling contact is also capable of instantaneous changes in direction.
Indeed, the BallBot \cite{lauwers2006dynamically,nagarajan2014ballbot} is a popular teaching tool and hobby project that leverages omni-directionality and radial symmetry by actuating a center sphere.
However, the BallBot requires complex control inputs to maintain stability due to its single point of contact with the ground, making it vulnerable to external perturbations and limiting its ability to traverse obstacles.

Here we present an untethered, radially symmetric legged robot configuration that is inherently stable and simple to control.
We leverage the energy efficiency of a rolling or wheeled robot while incorporating obstacle traversal abilities of a legged robot by designing a hybrid end effector with interchangeable rolling and high-friction contacts.
This form factor enables tripedal robots to perform straight-line and momentum-driven locomotion, and allows more rapid changes in direction than are possible with traditional quadrupedal robots, which facilitates more efficient navigation of cluttered/tight indoor spaces. 
Our design greatly enhances the maneuverability and stability of tripedal robots, enhancing their ability to  efficiently navigate cluttered, indoor spaces.
Our low-cost, simple manufacturing techniques allow the robot to be easily constructed and customized for a variety of applications, including teaching, delivery, and mapping.

In Section \ref{sec:design} we describe the design and construction of \robotname\ by combining a tripedal body form with a central rotating sphere.
In Section \ref{sec:results} we describe several unique maneuvers afforded by this novel robot configuration and present data on \robotname's velocity and turning radius.
Finally, we discuss additional gaits, functionalities, and practical applications in Section \ref{sec:discussion}.
Overall, this robot demonstrates how novel functionalities can arise from unique body forms, highlighting the importance of exploring beyond vertebrates for robotic design inspiration.

\section{Robot Design}\label{sec:design}

\subsection{Mechanical Design} \label{sec:prototype}

\begin{figure}[h]
    \centering
    \includegraphics[width=8cm]{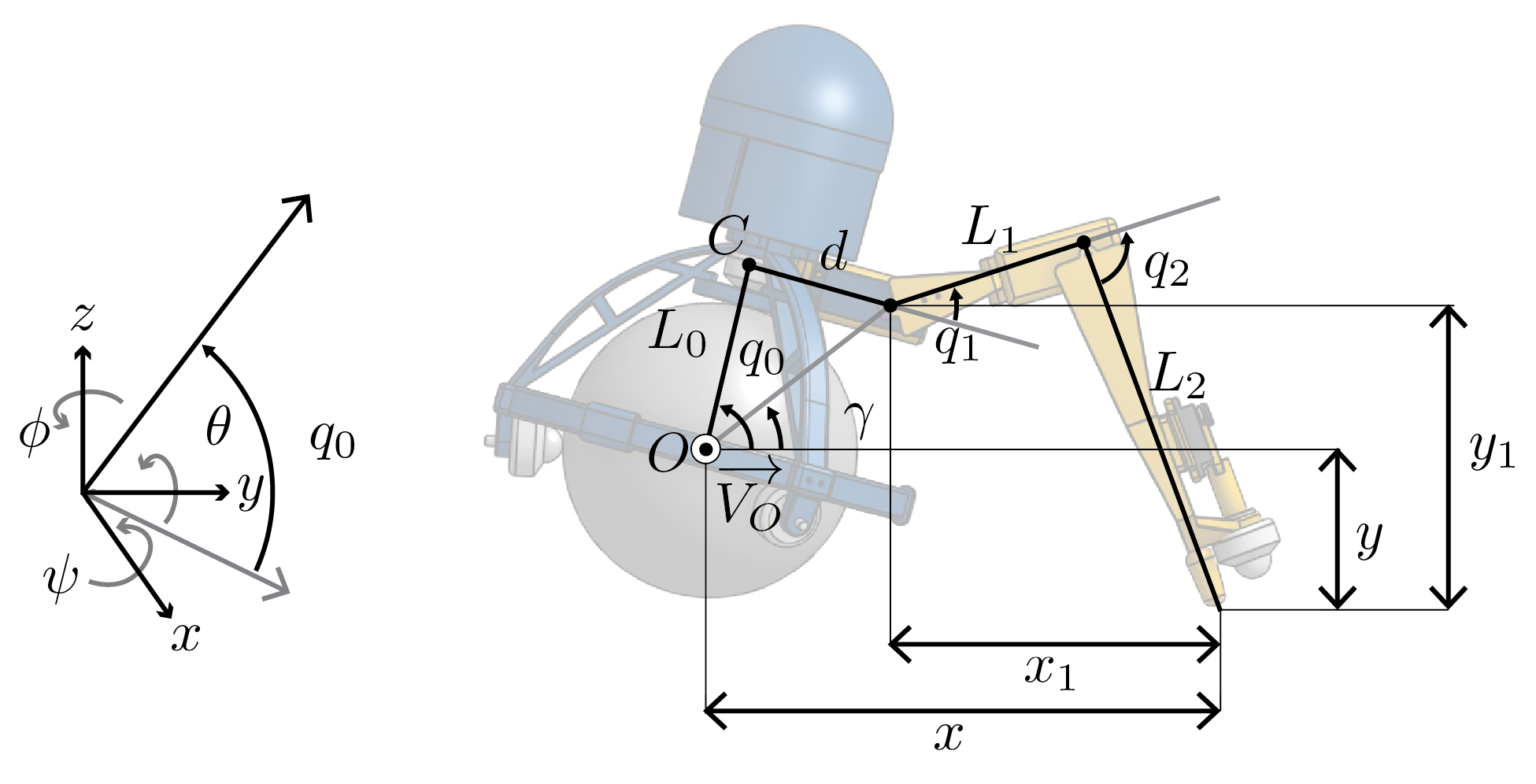}
    \caption{Kinematic chain for a single leg of \robotname.
    }
    \label{fig:sideview}
    \vspace{-.2cm}
\end{figure}

\robotname\ is held together by a central structure (Fig. \ref{fig:sideview}, shown in blue), consisting of a ``cage'' that encloses the ball and a ``control hub,'' which houses the electronics and battery.
The control hub which must be placed above the geometric center of the robot to maintain radial symmetry.
This positioning of the control hub aids the robot in maintaining stability in multiple configurations.
For example, when the robot tilts during locomotion, the mass of the control hub and the electronics therein shift so that the center of mass remains within the triangular base of support \cite{cartmill2002support}.

The central structure houses a larger center sphere (shown in gray) through four points of contact with spherical bearings, which allows for smooth motion in any direction.
This center sphere is used as an additional contact point during some gaits.
Additionally, the size of the center sphere determines the elevation of the axis of rotation for the central structure; a larger sphere increases stability because it reduces the effort to recover to an upright position.
A larger diameter sphere also makes the robot more robust to rolling on imperfection.
The center sphere's 200~mm diameter approaches the upper dimensional limit for most common commercially available 3D printers.
3D printing offers maximal control over the weight, size, and mass distribution of the sphere.

Three identical legs extend from the central structure.
Each leg includes two planar rotational joints: this configuration simplifies inverse kinematics while allowing the leg to reach around the center sphere. 
The flexion and extension of each leg joint is actuated by two commercially available servo motors {DS5160SSG180, Dongguan City Dsservo Technology Co, Dongguan, China) with a 180\textdegree~working angle. 
These servos were chosen for ease of control and affordability.

Additionally, at the end of each leg, a prismatic mechanism comprised of a smaller servo (MS24-F, Miuzei, Shenzhen, China) and a two-bar linkage extends and retracts a spherical bearing. 
We refer to this mechanism as a ``hybrid end effector'' (Fig. \ref{fig:overview} b}).
With the bearing extended, the leg has a rolling contact.
Retracting the bearing reveals a foot with a rubber cap to increase friction.
Henceforth, this frictional contact will be referred to as a ``foot.''
This hybrid end effector is key to enhancing \robotname's\ locomotion compared to other tripedal robots by allowing some legs to glide with minimal friction while other legs maintain stabilizing points of contact for the center structure.

The servos are powered by a lithium polymer battery with a nominal voltage of 7.4V, and controlled by an Arduino Uno microcontroller, both of which are housed in the central structure.
An MPU-6050 inertial measurement unit (IMU) (HiLetgo, Shenzhen, China) measures the robot accelerations and calculates its orientation.

Other than electronic components, fasteners, bearings, and bushings, all parts were 3D-printed in polylactic acid (PLA) using Filament Deposition Modeling (Ender 5, Creatily, Shenzhen, China).
The CAD files for the 3D printed parts, bill of materials, and assembly guide are available to download at our \deepblue (temporary link while under review, permanent doi will be updated upon acceptance).
The estimated cost for constructing the robot is approximately \$500 USD.

\subsection{Controller Design}\label{sec:controller}

To control \robotname, an inverse kinematic model has been solved analytically. 
Our inverse kinematic model works on only one ``active'' leg at a time and operates on the assumption that the active leg has three planar rotational degrees of freedom: two for the joints in the leg and one for the tilt that the entire robot (excluding the sphere) experiences about the vector normal to the plane of the leg and running through the centroid of the center sphere, $\overrightarrow{V_{O}}$.
To solve this model, we first solve for the degree of freedom $q_0$ corresponding to the robot's rotation about $\overrightarrow{V_{O}}$.
We can then use the result to convert our model into one with just two planar rotational degrees of freedom, $q_1$ and $q_2$.

To solve for $q_0$, we start from the center point of the center sphere $O$ and perform three successive transformations.
We first perform two rotational transformations corresponding to the pitch ($\theta$) and yaw ($\psi$) values measured from the onboard IMU, and subsequently one translation to the point $C$ in the robot's central structure, which is planar with all three upper leg joints.
We then solve for the angle of tilt with the vertical axis $q_0$, which we define as the vector between the point of contact of the center sphere with the ground and the center of the center sphere. 
The result of these calculations is the following:
\begin{equation}\label{q_0}    
    q_0 = \cos^{-1}(\frac{\sqrt{(L_0\cdot\cos{\psi}\cdot\sin{\theta})^2 + (L_0\cdot\sin{\psi})^2}}{L_0}) 
\end{equation}

Next, taking $x$ and $y$ as inputs, we solve for $q_1$ and $q_2$ as follows.
\begin{equation}\label{gamma}
    \gamma = q_0 - \tan^{-1}(\frac{d}{L_0})
\end{equation}

\begin{equation}\label{x_1}
    x_1 = x - \cos(\gamma)\cdot\sqrt{L_0^{2} + d^{2}}
\end{equation}

\begin{equation}\label{y_1}
y_1 = y - \sin(\gamma)\cdot\sqrt{L_0^{2} + d^{2}}
\end{equation}

\begin{equation}\label{q_2}
q_2 = \frac{\cos^{-1}(x_1^{2} + y_1^{2} - L_1^{2} - L_2^{2})}{2 \cdot L_1 \cdot L_2}
\end{equation}

\begin{equation}\label{q_1}
    \begin{split}
        & q_1 = 90 - q_0 - \tan^{-1}(\frac{y_1}{x_1}) \\
        & \hspace{0.7cm}+ \tan^{-1}(\frac{L_2 \cdot\sin(q_2)}{L_1+L_2\cdot\cos(q_2)})
    \end{split}
\end{equation}


Note that $q_2$ will always be negative with this solution, inducing an elbow-up joint configuration. 
We choose to exclusively use elbow-up configurations for all of our gaits.

To achieve various forms of locomotion on level surfaces, we leverage this inverse dynamics controller coupled with a series of mode switches between our foot and rolling contact. 
Additional details on these gaits are discussed in Section \ref{sec:dynamic}.

To control the physical prototype, we developed custom Arduino code that implements the inverse kinematic equations into motion controllers for varying gaits.
A simplified dynamical simulation of the robot was also generated using PyBullet and a simplified URDF model of the robot.
These are available at \deepblue.

\section{Locomotion Results}\label{sec:results}



\subsection{Maneuvers arising from hybrid end effector}\label{sec:static}

Actuation of the \robotname\ hybrid end effector is sufficient to perform several maneuvers without additional motion of the leg joints.
\robotname\ can execute all of these maneuvers with or without the center sphere.
It can ``stand'' with all rolling contacts to glide under external forces, such as a push or kick.
With all feet in the lowered position, it can ``brake'', which limits movement under external forces.
If it has momentum, \robotname\ can also perform a ``pivot'' maneuver by lowering one foot while the other two legs have rolling contacts (Fig. \ref{fig:turn}).
By controlling the duration of contact, the robot can pivot about the stationary foot to quickly rotate to arbitrary angles, freeing it from the 120\textdegree~turning constraint of other tripedal robots.
The turning radius is controlled by changing the distance between the center of the sphere to the stationary leg. These maneuvers can all be performed on level or inclined surfaces.

On inclined surfaces, \robotname\ can efficiently roll without any actuator motion.
It can also stand and brake in an ``orientation aware'' way by keeping the robot frame level (i.e., the control hub directly above the center sphere) while maintaining contact with all three leg end effectors.
Using this standing method, the robot gains additional stability on inclines.
This functionality likely facilitates more advanced maneuvers on inclines, such as uphill and transverse locomotion, because the position of the ground with respect to each end effector is known.


\subsection{Novel Gaits}\label{sec:dynamic}

\begin{figure}[h]
    \centering
    \includegraphics[width=\columnwidth]{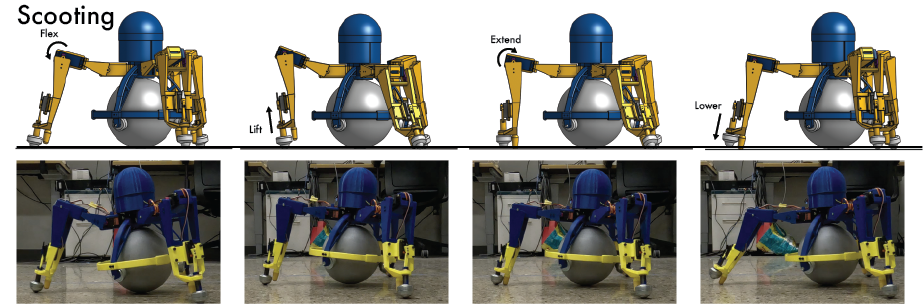}
    \caption{
    Sim-to-Real execution of scooting in simulation (top) and on a real robot (bottom).
    Scooting is quasi-static and retains all four points of contact throughout the gait cycle: 3 legs and the center sphere.
    The active leg can either push (forwards) or pull (backwards) with the frictional foot, then rolls in the recovery stage, while the rest of the robot remains rigid and moves along rolling contacts.}
    \label{fig:scoot}
    \vspace{-.2cm}
\end{figure}

The leg servos in conjunction with the hybrid end effector afford several locomotor gaits that generate movement with consistent headings on level ground.
The ``scooting'' gait uses one active leg that transitions back and forth between the rolling and foot contact to push itself in a straight-line trajectory, while keeping all three end effectors on the ground and the orientation of the robot constant and parallel to the ground (Fig. \ref{fig:scoot}). 
The two other (passive) legs lead the motion of the robot but maintain the same position throughout the gait.
To maintain stability throughout this motion, the robot must rotate slightly about the axis running through the contact point of the center sphere and the ground and parallel to the axes of rotation of the active leg.
This rotation creates a moment of force about the specified axis that prevents the robot from toppling, and allows the active leg to move throughout its reachable workspace without causing the robot to lose stability.

\begin{figure}[h]
    \centering
    \includegraphics[width=\columnwidth]{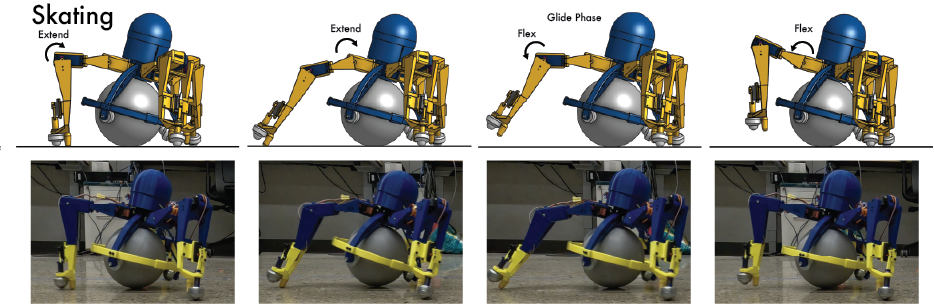}
    \caption{
    Sim-to-Real execution of skating in simulation (top) and on a real robot (bottom).
    With skating, the robot retains rolling contact via the center sphere and two wheels with rolling contacts.
    The active leg maintains the frictional foot mode throughout the entire stride.
    The frictional foot either pushes (forward) or pulls (backward) to propel the robot, then lifts off the substrate to return back to its starting position.
    }
    \label{fig:skate}
    \vspace{-.2cm}
\end{figure}

``Skating'' is a similar straight-line gait, but during this gait the robot lifts its active leg off of the ground between pushes, thereby eliminating the need to transition the state of the end effector of the active leg (Fig. \ref{fig:skate}). 
This gait is also designed to be more dynamic than scooting, as the leg lift allows for a rapid leg push that causes the center sphere to continue to roll after the duration of the push has terminated. 
This allows \robotname\ to increase speed over the course of multiple strides.

\begin{figure}[h]
    \centering
    \includegraphics[width=\columnwidth]{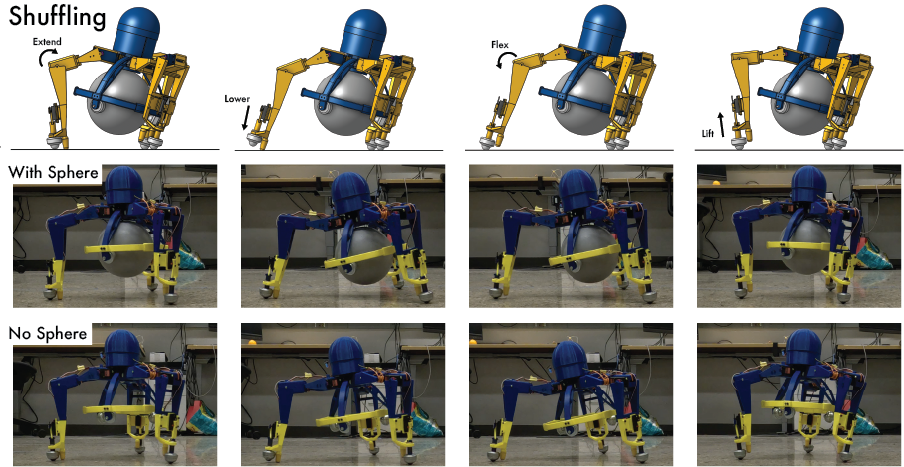}
    \caption{
    Sim-to-Real execution of shuffling in simulation (top) and on a real robot (bottom).
    With the center sphere lifted or removed, it can perform shuffling, which is equivalent to scooting, but with only the three legs contacting the ground.}
    \label{fig:shuffle}
    \vspace{-.2cm}
\end{figure}

``Shuffling'' locomotion is similar to the ``scooting'' gait, except that the ball is lifted in the air or absent during locomotion (Fig. \ref{fig:shuffle}).
In this gait, the only points of contact are the end effectors.
Each of these gaits can also be performed ``backwards'' with the one active leg ``pulling'' instead of pushing.
However, backwards gaits are not as efficient or fast as their forwards counterparts.

\begin{figure*}[h]
    \centering
    \includegraphics[width=\textwidth]{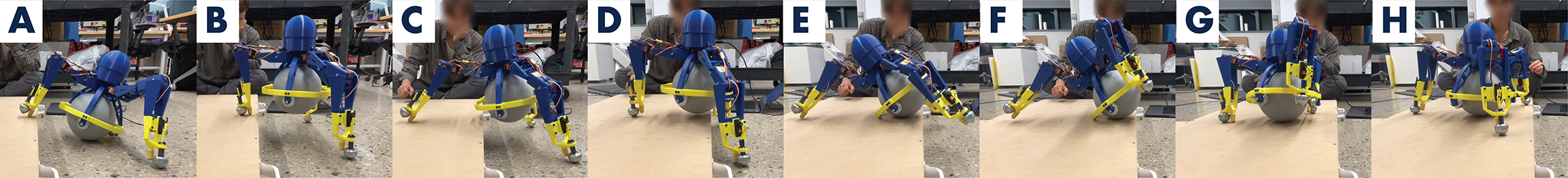}
    \caption{\robotname\ is capable of climbing stairs, thanks to its unique base of support configurations.
    In particular, panel F demonstrates how the robot can balance on only the center sphere and the central cage structure, to better reposition its legs and clear larger obstacles.}
    \label{fig:stairs}
\end{figure*}

\subsection{Tasks arising from linking maneuvers and gaits}\label{sec:links}

\robotname\ can instantaneously change heading without changing the orientation of the body by simply switching to a different leading leg (Fig. \ref{fig:turn}).

The robot can also traverse small obstacles by executing a sequence of multiple gaits and maneuvers.
This is demonstrated by traversing an electrical cord, which would prevent many small, wheeled delivery robots from operating in office environments (Fig. \ref{fig:overview}f).
\robotname\ first executes a backwards skate to lift the leading leg over the cord.
Then it performs a backwards shuffle to lift and translate the center sphere over and past the cord.
Finally, it performs two changes in direction using forward skating strides with each of the trailing legs to sequentially lift them over the cord.

Furthermore, the robot can climb up stairs that have a tread at least as deep as the diameter of its operating circle (Fig. \ref{fig:stairs}). 
This is demonstrated on a stair that has a rise height of 0.1~m, which is approximately 50\% of \robotname's hip height.
\robotname\ first places a single front foot on the stair, and keeping this foot stationary, pulls its center of mass towards the stair (Fig. \ref{fig:stairs}A).
Then the two back legs switch to foot contacts and extend until the center sphere clears the height of the stair (Fig. \ref{fig:stairs}B).
Next, the robot extends its rolling contact on the front leg to extend this contact point away from the center of mass (Fig. \ref{fig:stairs}C), and then switches back to the foot to pull the center ball fully onto the stair (Fig. \ref{fig:stairs}D).
At this point, the front foot switches to a rolling contact and extends out and upwards to induce a forward rotation of the robot about the central sphere (Fig. \ref{fig:stairs}E), causing the robot to shift to its sphere and cage base of support configuration (Fig. \ref{fig:overview}a).
Finally, the robot lifts its back legs above the surface of the stair (Fig. \ref{fig:stairs}F), and uses its front leg with its frictional foot mode to pull itself fully onto the stair (Fig. \ref{fig:stairs}G).

\subsection{Gait velocity and turning radius characterization}
In this section we present results from a set of experiments to characterize both the maximum speed of \robotname's various gaits and also calculate the turning radius for multiple direction-changing maneuvers.

\begin{table}[!tb]
    \centering
    \caption{The maximum velocities of various legged robots in meter per second (m/s) and Body Lengths per second (BL/s).
    Although two additional tripods achieved locomotion in physical robots, the speed was not reported \cite{lyons_rotopod_2007,ishikawa2012tripedal}.}
    \begin{tabular}{|c|c|c|}
    \hline 
    \textbf{Robot (Gait)} & Max Speed (m/s) & Max Speed (BL/s) \\ \hline
    MVA Tripod \cite{kim2019simple} & 0.01 & 0.11 \\ \hline
    \robotname (Scooting) & 0.16 & 0.72 \\ \hline
    STRIDER \cite{heaston_strider_2007} & 0.21 & 0.11 \\ \hline
    \robotname (Shuffling) & 0.39 & 1.74 \\ \hline
    \robotname (Skating) & 0.56 & \textbf{2.49} \\ \hline
    Ballbot \cite{nagarajan2014ballbot} & 0.75 & -- \\ \hline
    Spot \cite{SPOT} & 1.60 & 1.45 \\ \hline
    Ascento \cite{ascento2019} & 2.22 & -- \\ \hline
    \end{tabular}
    \label{table: max_velocity}
\end{table}

Table \ref{table: max_velocity} shows results for the maximum speed achieved by each of \robotname's gaits in both absolute speed and speed normalized by body length.
We also compare the maximum speed of \robotname's gaits to legged robots \cite{heaston_strider_2007, SPOT}, rolling robots \cite{nagarajan2014ballbot}, and robots with wheels on the ends of legs \cite{ascento2019}.
We find that skating is the fastest \robotname\ gait, yielding the highest normalized velocity across the set of examined robots. 

As stated in Section \ref{sec:intro}, the minimum turning radius determines maneuverability.
Table \ref{table: max_velocity} shows results for the minimum turning radius achieved by some of \robotname's turning maneuvers compared with other mobile robots.
Notice that both Ballbot and \robotname's turning without rotation maneuvers share a turning radius of 0 because even when traveling at their maximum speeds, these robots can instantaneously change heading without needing to change orientation (See Fig. \ref{fig:turn}).
When executing the pivot gait, \robotname\ required a minimum turning radius of 0.15m---a tighter turn than all other robots with legs considered.

\section{Discussion}\label{sec:discussion}


We present a novel form factor for an omni-directional legged robot that combines rolling and stepping in a radially symmetric configuration.
We characterize several new gaits afforded by the unique form factor and the hybrid rolling and frictional end effector.
This chimeric form factor achieves faster speeds and more consistent headings than traditional tripedal robots, tighter turns than bilaterally symmetric robots with wheels, and more stability and obstacle traversal capability than traditional ballbots.

Indeed, we argue that the stability and the versatility of the robot is due to the robust and unique form factor. 
With simple kinematic control and without any exteroceptive sensors, this robot is capable of object traversal, stair climbing, and recovery from lateral perturbations, all while maintaining dynamic and static stability.
This robustness makes \robotname\ suitable for mechanically mediated maneuvering, in which cluttered spaces can be navigated at high speeds by rebounding from collisions.
Mechanically mediated maneuvering is a successful strategy exhibited by small animals and robots with minimal feedback control \cite{jayaram2018transition, kovavc2009miniature, briod2014collision, klaptocz2010indoor, mintchev2017insect}.

\begin{table}[!tb]
    \centering
    \caption{The maximum velocities of various robots in meter per second (m/s) and Body Lengths per second (BL/s).
    Note that all robots with rolling contact have a tighter turning radius than the legged Spot robot.}
    \begin{tabular}{|c|c|c|}
    \hline 
    \textbf{Robot (Maneuver)} & Turning Radius (m) \\ \hline
    \robotname (Without rotation) & 0 \\ \hline
    \robotname (Pivot) & 0.153 \\ \hline
    Ballbot \cite{nagarajan2014ballbot} & 0 \\ \hline
    Ascento \cite{ascento2019} & 0.36 \\ \hline
    Spot \cite{SPOT} & 0.95 \\ \hline
    \end{tabular}
    \label{table: turning radius}
\end{table}


Additionally, due to its low construction cost and ubiquity of manufacturing using 3D printing, this robot has great potential as an educational tool.
Both the CAD models and the code for control are made freely available to facilitate widespread adoption and customization.


\robotname\ can easily be modified to greatly expand the range of tasks that can be performed with the robot.
Adding leg actuators capable of ab/adduction about the vertical axis running through the center of the control hub would increase maneuverability by facilitating the pivoting motion from a stationary position.
With more degrees of freedom and longer legs, \robotname\ would be capable of traversing greater obstacles, such as steeper stairs, tables, or large boulders.
This body plan could also accommodate additional appendages to increase stability, enable turning via translation in additional directions, or to use as manipulator arms.
However, it should be noted that additional degrees of freedom would greatly increase the control effort.

Future work with this robot entails adding a set of exteroceptive sensors to enable more autonomous localization, motion planning, and mapping.
With these incorporated, \robotname\ will be able to navigate cluttered indoor environments and surmount small obstacles, which is a limiting factor in the performance of current wheeled indoor robot platforms.
Example applications for \robotname\ include indoor exploration of cluttered and/or unstructured environments.
Another benefit of this form factor is that increasing payload increases stability, which makes this a particularly promising platform for a delivery robot.

\robotname\ has immense potential for navigating cluttered, indoor environments, such as in mapping and exploration, payload delivery, and teaching.

\section*{Acknowledgment}
The authors would like to thank Jessi Carlson for assistance with data collection.

\bibliographystyle{ieeetr}
\bibliography{ref.bib}

\end{document}